# Use of Dempster-Shafer Conflict Metric to Detect Interpretation Inconsistency

Jennifer Carlson and Robin R. Murphy[*]
Safety Security Rescue Research Center
University of South Florida
Tampa, FL 33620

## Abstract

A model of the world built from sensor data may be incorrect even if the sensors are functioning correctly. Possible causes include the use of inappropriate sensors (e.g. a laser looking through glass walls), sensor inaccuracies accumulate (e.g. localization errors), the *a priori* models are wrong, or the internal representation does not match the world (e.g. a static occupancy grid used with dynamically moving objects). We are interested in the case where the constructed model of the world is flawed, but there is no access to the ground truth that would allow the system to see the discrepancy, such as a robot entering an unknown environment. This paper considers the problem of determining when something is wrong using only the sensor data used to construct the world model. It proposes 11 interpretation inconsistency indicators based on the Dempster-Shafer conflict metric, *Con*, and evaluates these indicators according to three criteria: ability to distinguish true inconsistency from sensor noise (classification), estimate the magnitude of discrepancies (estimation), and determine the source(s) (if any) of sensing problems in the environment (isolation). The evaluation is conducted using data from a mobile robot with sonar and laser range sensors navigating indoor environments under controlled conditions. The evaluation shows that the Gambino indicator performed best in terms of estimation (at best 0.77 correlation), isolation, and classification of the sensing situation as *degraded* (7% false negative rate) or *normal* (0% false positive rate). While the evaluation is limited to 2D world models, the use of the Dempster-Shafer conflict metric to detect significant inconsistencies in data appears useful and could lead to systems which autonomously switch information sources to ensure the best mission performance, learn the relative contribution and reliability of sources for different environments, and reason about which sources to use under what circumstances.

## 1 INTRODUCTION

In an unknown environment, an intelligent agent, be it a human, software agent, or a robot, has only its senses (or sensors) to guide its actions. On the other hand, the suitability of each sense depends on the environment's characteristics (e.g. dark or noisy). If we assume that the agent does not know these characteristics, how can it determine which sensors to use and/or how to adapt when the sensing situation changes? This study addresses the problem of determining when an artificial agent's ability to sense the environment is impaired due to the use of inappropriate sensors, employing only the sensor data used to construct its world model.

This study considers the case of a mobile robot navigating in a controlled, but unknown, indoor environment in the presence of sensing anomalies. In this paper a *sensing anomaly* refers to *cases in which the physical sensor(s) are working within the manufacturer's specifications but the readings would lead to an incorrect interpretation of the environment*. For example, a sonar sensor has difficulty detecting smooth walls due to specular reflection, and laser range scanners cannot detect (clean) glass surfaces. The presence of sensing anomalies is an indication that the selected sensor, or set of sensors, are inappropriate for a given environment.

Features of the Dempster-Shafer weight of conflict metric, *Con*, are examined in this study as potential solutions to the problem of detecting the use of inappropriate sensors. The rationale for exploring Shafer's model of conflict (Shafer 1976) arises from previous work in the field of robots (see Section 2) and the assumption that all environments are consistent, therefore a metric that measures inconsistency could potentially detect sensing problems without relying on a ground truth. This potential was initially explored in (Carlson *et al.* 2005) where the *Con* metric was shown to provide an estimate (0.7–0.9 correlation) of the quality of the world model (a 2D map) for sonar readings in indoor and confined space environments. In this paper the performance of 11 distinct features were evaluated according to three criteria: the ability to estimate the state of a robot's sensing capabilities (estimation), distinguish true sensing anomalies from sensor noise (classification), and determine the source of the sensing problem within the environment (isolation).

[*]Email: {jcarlso1,murphy}@csee.usf.edu

The results of 30 experiments, using the real sonar and laser readings, showed that the Gambino indicator adapted from (Gambino, Ulivi, & Vendittelli 1997) could serve as a general solution to the problem of detecting, estimating, and isolating sensing problems in unknown environments. It was found to estimate the overall error in the occupancy grid (0.77 correlation at best) and classify the sensing situation with a 0% false positive and a 7% false negative rate. In addition it performed at least as well as the other indicators on the task of isolating problems within a 2D occupancy grid. The performance of the remaining 10 interpretation inconsistency indicators varied across the three tests. The *maximum increase*, *increase frequency* and *area* indicators in particular showed potential as specialized tools.

## 2  RELATED WORK

Inconsistency has been used as a tool in two overlapping fields relevant to this work: mobile robotics and fusion. In both fields the information used for action and decision making is often drawn from multiple, possibly unreliable and/or incompatible sources. The prevalence of formal models in these fields allows the degree of inconsistency to be measured.

In mobile robotics measures of inconsistency, such as conflict and entropy, have been used to solve the problem of localization in partially known or unknown environments. Examples of these can be found in (Ayrulu & Barshan 2002), (Baltzakis, Argyros, & Trahanias 2003), (Davidson & Murray 2002), and (Moreno & Dapena 2003). These approaches often rely on environment-specific assumptions, ranging from a ground truth map to loose constraints like "the ground plane is flat" to reduce complexity. They typically choose actions or make decisions which minimize inconsistency.

Three studies found in the robotics literature develop approaches for solving sensing problems using inconsistency. In (Gambino, Ulivi, & Vendittelli 1997) a metric was developed based on conflict (Smets' formulation (1990)) to determine when information becomes irrelevant in unknown dynamic environments. The results showed that this metric allowed the simulated robot to respond to changes in the environment faster as compared to the traditional Dempster-Shafer mapping approach. Yi *et al.* (2000) developed a metric to adapt their sonar sensor model using the conflict generated by the latest measurement. The resulting map, built using a Nomadics Super Scout II, was more internally consistent (fewer outliers) but not accurate compared to the ground truth. In (Shayer *et al.* 2002) a novel inconsistency metric is presented based on an adaptive fuzzy-logic sensor fusion technique developed in previous work. The metric compares each pair of sensors by creating a fused map from their readings. The agreement rate between the two sensors' original map and the fused map was used to provide a ranking of each individual sensor's relative accuracy. The sensors were correctly ranked 83% of the time.

An overview of the broad area of fusion that highlights the problem of inconsistency and various approaches for handling it, can be found in (Appriou *et al.* 2001). Related work along these lines can be found in (2001) where Ayoun and Smets present an approach to the association problem which selects the assignment of sensor readings to targets which minimizes the conflict (Smets formulation (1990)) of the fused information. The approach is validated by a series of simplified (static world, binary sensor output) examples and methods for generalizing the approach.

Dempster-Shafer theory was selected as a potential solution partly due to earlier work showing equal or better map building performance compared to Bayesian and fuzzy approaches when data from multiple sensors are fused. The analysis of evidential structures presented in (Zhu & Basir 2003) showed that Dempster-Shafer theory is equivalent to, and has the same positive characteristics as both Bayesian and fuzzy approaches. Pagac, Nebot, & Durrant-Whyte (1998) showed that Dempster-Shafer outperformed Bayesian updating methods for building occupancy grids from sensor readings.

## 3  APPROACH

The approach taken in this study examines features of *Con* as potential solutions to the problem of de-tecting when sensors are inappropriate in unknown environments. Given the assumption that an environment is consistent, the Dempster-Shafer weight of conflict metric, *Con*, provides a means of measuring the degree of inconsistency and therefore the severity of sensing problems in unknown environments. Eleven *interpretation inconsistency indicators* were developed to explore the potential of this approach. A interpretation inconsistency indicator is defined as an *indicator that uses features of conflict to determine if observations and/or interpretations used to update a given model are* suspect, *i.e. should not be trusted, or* normal Three performance metrics, Fisher's Linear Discriminant (FLD), Pearson's correlation coefficient, and Baddeley's $\Delta^2$ metric were used to test the ability of each interpretation inconsistency indicator to determine when the current sensor(s) are inappropriate, measure the extent to which this is the case, and to isolate problem areas.

### 3.1 CONSISTENCY ASSUMPTION

It is presupposed that any environment will conform to the *consistency assumption*, for example a point in space cannot be both occupied and empty in the same instant. Drawing from logic (Ebbinghaus, Flum, & Thomas 1994), this assumption is modeled as the absence of conflict ($\phi$), i.e. $\psi \cap \neg\psi = \phi$, where $\psi$ is any hypothesis and $\phi$ must be the empty set.

Given this *consistency assumption*, if conflict appears in a sensor-based world model it may be caused by one or more of the following: sensor noise (e.g. introduced by discretization), the use of inappropriate sensors, inaccurate *a priori* models, or the use of a flawed internal representation (e.g. a static occupancy grid used for dynamically moving objects). Therefore, depending on the circumstances in which sensor readings are gathered and interpreted, conflict can theoretically be used to detect a variety of sensing problems.

### 3.2 DEMPSTER-SHAFER CONFLICT

The Dempster-Shafer weight of conflict metric, *Con* measures the support for conflicting evidence in a set of observations, thus given the *consistency assumption*, this metric can be applied to detect sensing problems in a world model. The metric itself is given in Equation (1)

$$
\begin{aligned}
Con(Bel_1, Bel_2) &= log(\frac{1}{1-k}) \quad (1) \\
\text{where} \\
k &= \sum_{A_i \cap B_j = \emptyset} m_1(A_i) m_2(B_j)
\end{aligned}
$$

where $Bel_1$ and $Bel_2$ represent degrees of belief for the sets of hypotheses $\theta_1$ and $\theta_2$ respectively, $A_i \in \theta_1$, $B_j \in \theta_2$, and $m_1(x)$ and $m_2(x)$ give the belief assigned to a hypothesis $x$ by $Bel_1$ and $Bel_2$ respectively. *Con* takes a value between 0 and $\infty$; as $k \rightarrow 0.0$, $Con \rightarrow 0.0$, and as $k \rightarrow 1.0$, $Con \rightarrow \infty$. It is additive, which means that the conflict from more than two belief functions can be attained easily.

This study also examines features of Smets' *conflict* belief mass as potential solutions to the problem of detecting when sensors are inappropriate in unknown environments. In Smets' model (1990) $\phi$ is treated as a valid hypothesis that accumulates belief, rather than an error to be factored out. It remains to be seen which of these two models of conflict is more effective for solving the problem of detecting sensing problems.

### 3.3 EVALUATION

The interpretation inconsistency indicators were evaluated using three tests. Each test used a *quantitative map quality metric* as a measure of the actual sensing situation. In this study a traditional occupancy grid (Elfes 1989) was used. This divides a two dimensional space into a set of equally sized cells, each labeled occupied or empty with some level of certainty (probabilistic (Thrun 2003), possibilistic (López-Sánchez, de Mántaras, & Sierra 1997), and evidential models (Murphy 2000) have been used with such grids in mobile robot mapping). The quality of an occupancy grid given a ground truth occupancy grid can be quantified using Equations (2–4),

$$
\begin{aligned}
Error &= \sum_{x=0}^{w}\sum_{y=0}^{h} MAX(error_o, error_e) \quad (2) \\
error_o &= |grid_o(x,y) - truth_o(x,y)| \quad (3) \\
error_e &= |grid_e(x,y) - truth_e(x,y)| \quad (4)
\end{aligned}
$$

where *width* and *height* together give the size of the occupancy and ground truth grids, $grid_o(x,y)$, $grid_e(x,y)$ give the occupancy and empty values from the occupancy grid, and $truth_o(x,y)$ and $truth_e(x,y)$ the same values from the ground truth grid, respectively. Lower *Error* scores indicate a better match with the ground truth.[1]

The ability to estimate the overall map quality, i.e. the extent to which a robot's current sensing is inappropriate, was measured using Pearson's correlation coefficient (Dowdy, Weardon, & Chilko 2004). In this case the test was used to determine if the average value assigned by an interpretation inconsistency indicator over an occupancy grid (see Section 4.1) varied linearly with the grid's *Error* score.

---
[1] Any cells which were not considered in the ground truth (for example cells on the other side of walls), or not yet scanned by the sensors were excluded from the overall error calculation.

Fisher's linear discriminant (FLD) (Fisher 1936) was used to test each interpretation inconsistency indicator's ability to correctly detect a situation in which the current sensor(s) are inappropriate. The two-class formulation of FLD, given in Equation (5) where $\mu$ is the mean value assigned to a class of grids (accurate versus inaccurate in this case) and $\sigma$ is the variance within each group, provides a straightforward means of testing the separability of classes provided by an indicator.

$$FLD = \frac{|\mu_{class\ 1} - \mu_{class\ 2}|^2}{\sigma_{class\ 1} + \sigma_{class\ 2}} \quad (5)$$

Before the classification test (FLD) could be applied, the ground truth classification of the occupancy grids from each run had to be determined. The *k-means* clustering technique (Hartigan & Wong 1979) was used to find sets of *Error* scores that were statistically close in value. This way three distinct sets were identified, the largest of which contained error values subjectively determined to reflect accurate occupancy grids, with the other two reflecting inaccurate grids. A threshold value of 300 was selected that lay between these two groups.

The third test applied the $\Delta^2$ metric (Baddeley 1992) to report how well an interpretation inconsistency indicator can isolate sensing problems. The $\Delta^2$ metric is a position-sensitive (but shape insensitive) metric for evaluating binary images produced by computer vision algorithms. It applies Equation (6) to a pair of images (typically the estimate image $B$ and the ground truth image $A$), i.e. $\Delta^2(A, B) =$

$$\sqrt{\frac{1}{N} \sum_{x \in X} (min(d(x, A), c) - min(d(x, B), c))^2} \quad (6)$$

where $X$ is the set of highlighted pixels in either image, $d(x, I)$ corresponds to the distance between a pixel $x$ and the nearest highlighted pixel in image $I$, and $c$ is an arbitrary constant (in this study $c = 100$). $\Delta^2$ produces a single scalar measure which describes how well the two images match in terms of both false negatives and false positives.

## 4 IMPLEMENTATION

A series of 30 experiments using sonar or laser data from a Nomad 200 robot were performed to determine how well each of the interpretation inconsistency indicators could estimate the overall map quality, classify the sensing situation, and isolate difficult to sense regions within the environment. The indicators applied threshold values to determine which cells were suspicious and which were considered normal. Those threshold values were varied for each indicator resulting in the 355 potential indicators examined in this study. Fifteen runs in three indoor hallways (five each) were evaluated using an indicator and the map quality metric. This process was repeated individually for the sonar and the laser readings, resulting in 30 experiments used to measure each of the 355 combinations' capabilities. The experimental procedure used controlled environments to ensure that the use of inappropriate sensors would generate significantly more conflict than sensor modeling or representation errors (see Section 3.1).

### 4.1 INTERPRETATION INCONSISTENCY INDICATORS

Eleven different interpretation inconsistency indicators were developed using either Shafer's *Con* metric or Smets' conflict belief mass (both of which will be referred to as *conflict*, except where the distinction is important). The indicators labeled each cell in the grid as *suspect*, if the conflict indicated a problem with sensing, or *normal* otherwise. The label was chosen on a cell by cell basis (with the exception of the *area* indicator, see Table 1) by thresholding a function of the conflict value. The average of this function's value (after thresholding) over the entire grid is called the *conflict score* for that indicator. In addition each indicator produces a *conflict map* which is a binary image in which *suspect* cells are labeled *true* and the rest *false*.

Table 1 lists each indicator along with the set of threshold values tested for that indicator. In the analysis the thresholds were increased in constant intervals, given in the Step column, from the minimum to the maximum value shown in the Threshold Range column. Note that in this first exploratory study the appropriate threshold values to test were determined empirically.

The 11 indicators can be characterized by describing the function that each applied to the conflict prior to thresholding. The *total* indicator takes the sum of the *Con* metric generated by each update, i.e. Shafer's formulation. Three additional indicators normalize *total*'s function prior to thresholding according to the current sensor's *angular resolution*, *range resolution*, and average *update rate* respectively. *Maximum increase* takes the maximum value of *Con* achieved in a single update. The *average* and *average sequence* in-dicators both use the average $Con$ generated per update, with the latter restricted to the last contiguous set of conflicting updates. *Frequency* and *increase frequency* both consider the percentage of updates that generated conflict, where *frequency* counts all updates with $Con > 0$ but *increase frequency* counts only those where $Con \geq magnitude$. The Gambino indicator uses a heuristic developed in (Gambino, Ulivi, & Vendittelli 1997): if either *occupied* or *empty* were at least 50% and Smets' conflict belief mass increased by at least 10% in a single reading then something is wrong. In this study the number of times in which these conditions held was counted and then thresholded. Finally the *area* indicator uses connected components to find contiguous regions in a conflict map given by a *total* indicator (using the first threshold listed in Table 1), and then applies the *size* threshold to remove smaller (assumed to be noise) regions.

Table 1: The interpretation inconsistency indicators evaluated in this study. Note that the *area* and *increase frequency* use a pair of threshold values to determine if a given cell is *suspect*.

| Interpretation inconsistency indicator | Threshold Range | Step |
|---|---|---|
| *area* | 0.25–5 | 0.25 |
|    size | 50–250 | 50 |
| *average* | 0.025–0.5 | 0.025 |
| *average sequence* | 0.05–1.0 | 0.05 |
| *frequency* | 0.05–0.95 | 0.05 |
| Gambino | 0.5–10.0 | 0.5 |
| *increase frequency* | 0.05–0.95 | 0.05 |
|    magnitude | 0.5–2.0 | 0.5 |
| *maximum increase* | 0.1–2.0 | 0.1 |
| *normalized by:* | | |
| angular resolution, | 0.025–0.5 | 0.025 |
| range resolution, | 0.25–5.0 | 0.25 |
| and update rate | 0.025–0.5 | 0.025 |
| *total* | 0.25–5.0 | 0.25 |

### 4.2 EXPERIMENTAL PROCEDURE

A Nomadic Technologies Nomad 200 robot (see Fig. 1(a)) was used to simultaneously collect sonar and laser readings in indoor hallways. Three different uncluttered hallways with the characteristics presented in Table 2 were selected. The robot was teleoperated down the center of the hallway for a distance of six meters while readings from the robot's single ring of 16 sonar sensors, and a Sick laser sensor mounted just above them, were collected for offline analysis. The first five good runs in each hallway, in terms of consistent sampling rates and the absence of errors in the data collection process, were used in this paper.

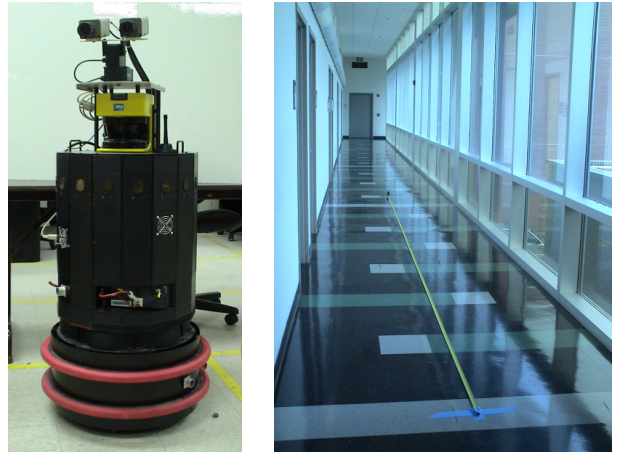

(a) Nomad 200 robot     (b) The *window* hallway

Figure 1: The robot used for data collection and one of the hallways used as a testbed environment.

Table 2: Characteristics of the three indoor hallways used to test the performance of the interpretation inconsistency indicators. The width (W) and length (L) are reported in meters. Smooth refers to painted sheetrock walls.

| Hallway | Sonar | Laser | W | L | Walls |
|---|---|---|---|---|---|
| *narrow* | poor | good | 1.8 | 11.2 | Smooth |
| *wide* | poor | good | 2.5 | 14.2 | Smooth |
| *window* | poor | poor | 2.0 | 27 | Smooth & large windows |

The offline analysis system used sensor models to register either the sonar or the laser readings to a 28 meter square occupancy grid, with a cell resolution of 10.16 cm (four inches) in both the $x$ and $y$ directions. This paper follows the approach described in (Murphy 2000) for building occupancy grid maps from sensor readings using Dempster-Shafer theory. In this approach sensor models are used to convert range readings, $d$, into belief masses, $m$, distributed over a cone. The cone has a radius of $R$ and a half-angle $\beta$ where $\beta$ is commonly tuned to the particular environment.

The Dempster-Shafer (or Smets) occupancy grid was updated using the sensor readings until the robot had traveled 1.0 meters, at which point the grid was evaluated using an interpretation inconsistency indicator and the quantitative map quality metric. (In order to use this metric, a ground truth occupancy map was manually created for each hallway.) This process was repeated every half meter, resulting in 10 sampled locations per experiment.

For each of the 15 runs this procedure was repeated for the sonar, then the laser data, resulting in 300 sampled locations (in 30 time series) recorded for use in a post-hoc analysis. The following three values were recorded for each point: the *Error* score, the conflict score, and the value given by the $\Delta^2$ metric which compared the conflict map generated by a given interpretation inconsistency indicator to an *error image*. The error image was produced by applying standard binarization techniques to the grayscale image produced from each cell's error, i.e. $MAX(error_o, error_e)$ from Equations (2–4). The post-hoc analysis generated Pearson's correlation coefficient, Fisher's Linear Discriminant, and the $\Delta^2$ score for each indicator and sampled location. The results of this analysis are presented in Section 5.

## 5 RESULTS

The performance of the 11 distinct interpretation inconsistency indicators were evaluated using the three performance metrics presented in Section 3.3. Both the average and the best performance of each indicator (across the threshold values tested) were used to evaluate that indicator's ability to perform each of the three tasks: classification, estimation, and isolation of sensing problems. In general, the interpretation inconsistency indicators showed varying performance across the three tests. *Area* and *increase frequency* for example ranked among the top three indicators on the estimation and isolation test, but achieved only average performance on the classification test. *Maximum increase* showed the inverse trend by achieving the best overall performance on the classification test while showing average performance on the other two tests. The Gambino indicator was the only indicator that performed well on all three. At the threshold value of 2.0 it was found to classify the sensing situation with a false negative rate of 7%, estimate the *Error* metric (0.77 correlation), and isolate sensing problems within the grid at least as well as the other indicators.

For each of the three tests the mean (average), variance ($\sigma^2$), and best performance (minimum or maximum depending on the metric) of each interpretation inconsistency indicator are presented in both tabular and graphical form. Also included are the number of threshold combinations that contributed to each indicator's results (see Table 1). (For example *area* used two thresholds to determine which cells were *suspect*. These were varied over 20 and five values respectively, resulting in 100 combinations tested.) In the graphs (Figures 2(b), 3(b), and 4(b)) the means are represented as a bar graph with variances shown as error bars. The best performance is plotted as a line graph along the same $y$ axis. For all three tests the results are presented alphabetically to facilitate comparison.

Table 2(a) and Figure 2(b) give the results of the estimation test. This test used Pearson's correlation coefficient which varies between $-1.0$ for a perfect negative linear relationship and 1.0 for a perfect positive linear relationship. Note that the conflict score is only bounded for the frequency and increase frequency indicators (i.e. a strong negative relationship is just as useful for estimation with these indicators as an equally strong positive relationship). All peak correlations were tested for statistical significance with the appropriate corrections for autocorrelation (Meko 2005) with and without Bonferroni correction (assuming 355 trails).

Table 2(a) and Figure 2(b) show that the majority of the interpretation inconsistency indicators provide poor estimates of *Error*. Only *area* and the Gambino indicator, at their peak performance, showed the ability to estimate the total error score for both sonar and laser readings, with the Gambino indicator performing best in terms of both average (mean) and best performance. This result is interesting when compared to previous results in indoor hallways (see Carlson *et al.* (2005)) which showed a very strong (0.92) correlation between the *total* indicator and the overall error score for sonar readings alone. It is clear from this result that a more sophisticated indicator was needed to handle readings from both sonar and laser sensors.

The estimation results also demonstrate the sensitivity of each interpretation inconsistency indicator to variance in its parameter(s). *Total* and its normalized variants appear to be insensitive ($\sigma^2 < 0.0004$) to changes in their respective threshold values. *Area* was the most sensitive ($\sigma^2 = 0.092$), with the Gambino indicator coming in second ($\sigma^2 = 0.090$). A detailed analysis of the raw correlation results shows that three of the indicators, *average sequence*, *average*, and the Gambino indicator, produced a monotonic increase or decrease (relative to the threshold value) from their best performance to near zero correlation. This suggests that these three indicators are good candidates for systems in which tuning (or learning) of suitable thresholds can be performed.

The isolation test used Baddeley's $\Delta^2$ metric to compare the error and conflict maps, the results of which are shown in Table 3(a) and Figure 3(b). For the $\Delta^2$ metric smaller values indicate a better ability to isolate sensor problems within the grid. This metric does not have a set bound like the correlation coefficient, thus it is only useful for comparing the performance of the 11 indicators.

The isolation results given in Table 3(a) showed little variance both within (at most 0.4) and between (0.70) the interpretation inconsistency indicators. This re-

| Indicator | Mean | $\sigma^2$ | Best | N |
|---|---|---|---|---|
| Angular resolution | -0.492 | 0.000 | -0.484 | 20 |
| Area | 0.090 | 0.092 | 0.677** | 100 |
| Average | -0.314 | 0.025 | 0.033 | 20 |
| Average sequence | -0.218 | 0.064 | 0.064 | 20 |
| Frequency | -0.041 | 0.026 | -0.265 | 19 |
| Gambino | 0.331 | 0.090 | 0.765* | 20 |
| Increase frequency | 0.033 | 0.048 | 0.412* | 35 |
| Maximum increase | -0.452 | 0.011 | -0.275 | 20 |
| Range resolution | -0.474 | 0.000 | -0.464 | 20 |
| Total | -0.474 | 0.000 | -0.464 | 20 |
| Update rate | -0.396 | 0.000 | -0.351 | 20 |

(a)

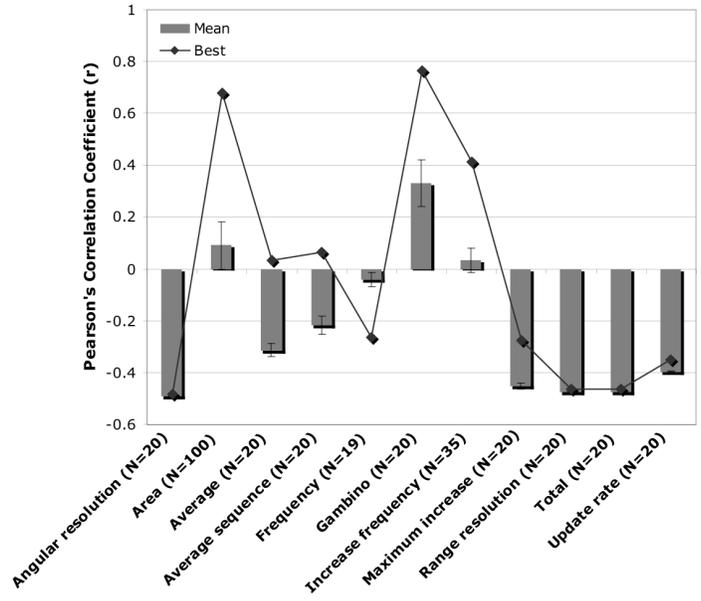

(b)

Figure 2: Results of the estimation test showing the correlation between the average conflict score and $Error$ for each of the 11 indicators. N is the number of thresholds values tested for each metric. Statistically significant correlations are mark with * ($\alpha$= 0.05) or ** (Bonferroni corrected $\alpha$= 1.41E-4).

| Indicator | Mean | $\sigma^2$ | Best | N |
|---|---|---|---|---|
| Angular resolution | 9.68 | 0.00 | 9.65 | 20 |
| Area | 7.95 | 0.40 | 7.39 | 100 |
| Average | 8.48 | 0.18 | 7.86 | 20 |
| Average sequence | 8.33 | 0.09 | 7.91 | 20 |
| Frequency | 8.84 | 0.18 | 8.04 | 19 |
| Gambino | 7.61 | 0.36 | 7.37 | 20 |
| Increase frequency | 7.59 | 0.19 | 7.37 | 76 |
| Maximum increase | 9.20 | 0.11 | 8.60 | 20 |
| Range resolution | 9.30 | 0.04 | 9.06 | 20 |
| Total | 9.30 | 0.04 | 9.06 | 20 |
| Update rate | 9.12 | 0.04 | 8.82 | 20 |

(a)

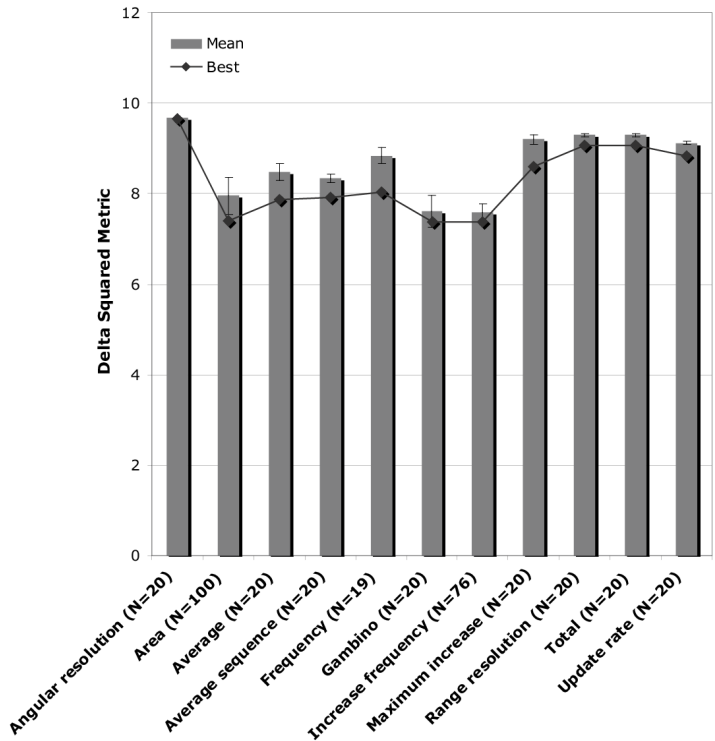

(b)

Figure 3: Results of the isolation test showing the $\Delta^2$ metric for each of the 11 indicators. N is the number of thresholds values tested for each metric.

sult is in part due to the small size of the grid and subsequently that of the error image and conflict map compared using the $\Delta^2$ metric. As with the estimation test, *total* and its normalized variants performed poorly compared to the other indicators and showed the least sensitivity to their threshold values. *Area* and the Gambino indicator again showed the most sentivity. Although the results of the estimation and isolation results are similar they are not identical. On the isolation test the *increase frequency* indicator performed marginally better than the Gambino indicator with half the variance.

The classification test results are given in Table 4(a) and Figure 4(b). This test was performed using Fisher's Linear Discriminant (FLD) which, like $\Delta^2$, does not vary within a set scale. In this case larger values indicate a better ability to classify the sensing situation.

The classification results in Table 4(a) showed that the interpretation inconsistency indicators were more varied in their ability to distinguish between normal and degraded sensing situations as compared to their isolation ability. Variance in performance within the interpretation inconsistency indicators was marginal, with the Gambino indicator and *maximum increase* showing the largest variances. Note that although the overall variance is low, for the majority of the indicators the difference between mean and best performance is noticeably different. In this test *total* and its normalized variants performed better (in a relative sense) then in the other two tests. They ranked highest in terms of mean or average behavior, with the Gambino indicator and *maximum increase* only performing better for a single threshold value.

Due to the Gambino indicator's performance on the estimation and isolation tests its ability to classify the sensing situation was examined further. For all maps with error scores below the 300 threshold, the Gambino indicator (at a threshold of 2.0) had an average conflict score of 0.0, indicating that not a single cell within the grid had been flagged as degraded. For maps with degraded error scores ($\geq 300$) only 7% also had scores of 0.0, while the rest had average conflict scores of 3.0 or more.

Overall the results show that good performance on one of the tests does not imply good performance on the other two. For example *maximum increase* ranked low on the estimation and isolation tests, but outperformed the the other indicators on classification. *Increase frequency* performed best on isolation, worst on classification, and estimated the *Error* metric moderately well (0.42 correlation). The *area* indicator ranked high on two of the tests: estimation (0.68 correlation) and isolation. The only interpretation inconsistency indicator that performed well on all three was the Gambino indicator which achieved the best overall performance on estimation, and second best on classification and isolation. This indicator's 0% false positive and 7% false negative rates also indicate that it would be a good candidate solution for the problem of detecting sensing problems in unknown environments.

# 6 CONCLUSIONS

The analysis of 11 interpretation inconsistency indicators, developed and tested using real sensor data from a mobile robot operating in controlled indoor environments, show that the Gambino indicator could serve as a general solution to the problem of detecting, estimating, and isolating sensing problems in unknown environments. A series of 30 experiments showed that the Gambino indicator, using a threshold of 2.0, could estimate the overall error in the occupancy grid (0.77 correlation), classify the sensing situation with a false negative rate of 7%, and perform at least as well as the other indicators on isolating problems within the grid. The analysis also shows that the *area*, *maximum increase*, and *increase frequency* indicators (which also performed well on at least one of these tasks) could be developed into specialized tools.

This exploratory study has developed a general approach to the problem of detecting sensing problems in unknown environments, relying only on the assumption of consistency and Dempster-Shafer theory. The experiments reported in this study represent one possible application of this approach, i.e. to detecting inconsistency in sensor readings with 2D models. More work is needed to explore its full potential to capture inconsistency in a variety of information sources and belief spaces.

One problem for any inconsistency-based approach like this one is the homogeneity of the sources, especially in terms of the conditions in which they will produce inappropriate (or simply incorrect) information. For applications like mapping in mobile robots, where only a few sources are used and the conditions (environment) change rapidly, this issue rarely appears. Alternate applications of this approach in which many sources (e.g. websites on the internet) are consulted and the information that is gathered is at a higher (e.g. symbolic or decision) level, homogeneity may be more of a problem. As potential applications of this approach are explored, the failure conditions of the sources and the sensitivity of a specific interpretation inconsistency indicator to the homogeneity of those conditions should be examined.

| Indicator | Mean | $\sigma^2$ | Best | N |
|---|---|---|---|---|
| Angular resolution | 9.94E-3 | 7.94E-8 | 1.03E-2 | 20 |
| Area | 1.71E-3 | 5.33E-6 | 9.29E-3 | 100 |
| Average | 1.72E-3 | 1.04E-6 | 2.97E-3 | 20 |
| Average sequence | 1.80E-3 | 3.26E-6 | 6.29E-3 | 20 |
| Frequency | 1.05E-3 | 1.59E-6 | 3.36E-3 | 19 |
| Gambino | 3.02E-3 | 2.95E-5 | 1.83E-2 | 20 |
| Increase frequency | 3.39E-4 | 1.45E-7 | 1.59E-3 | 35 |
| Maximum increase | 7.18E-3 | 2.31E-5 | 1.95E-2 | 20 |
| Range resolution | 8.84E-3 | 1.84E-7 | 1.00E-2 | 20 |
| Total | 8.84E-3 | 1.84E-7 | 1.00E-2 | 20 |
| Update rate | 6.30E-3 | 5.66E-7 | 7.29E-3 | 20 |

(a)

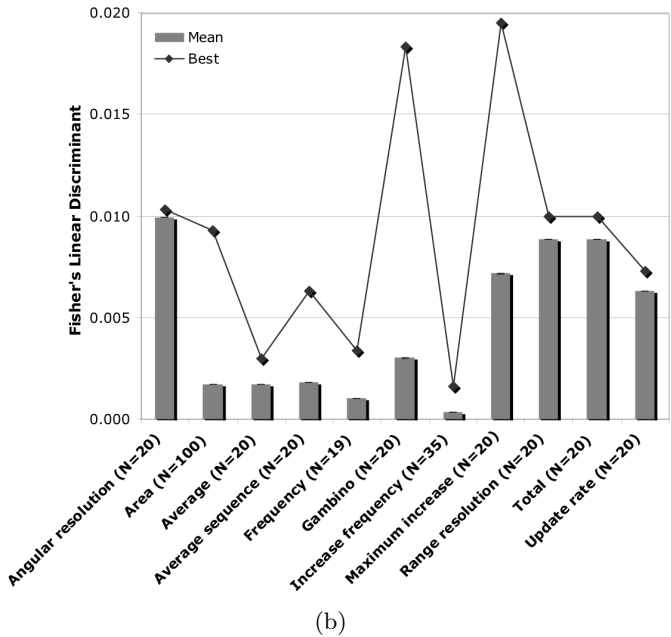

(b)

Figure 4: Results of the classification test showing Fisher's Linear Discriminant (FLD) for each of the 11 indicators. N is the number of thresholds values tested for each metric.

Despite this vulnerability the field of fusion, in addition to robotics, could benefit from the approach presented in this paper which treats inconsistency as a tool to evaluate the fused result. Such an approach could provide feedback for a fusion system, to help resolve the problems of association (determining which bits of information are describing the same thing) and reliability estimation described as important challenges in (Appriou et al. 2001). This could lead to systems which adaptively switch information sources to optimize mission performance, learn the relative contribution and reliability of sources for different environments, and reason about which information sources to use under what circumstances.

## Acknowledgments

This effort was supported by grants from the Office of Naval Research (N00773-99PI543), the Department of Energy (DE-FG02-01ER45904), and SAIC. The authors would like to thank Jennifer Burke and the anonymous reviewers for their helpful comments.